\documentclass[10pt,twocolumn,letterpaper]{article}

\usepackage{cvpr}              % To produce the CAMERA-READY version
\usepackage{graphicx}
\usepackage{amsmath}
\usepackage{amssymb}
\usepackage{booktabs}
\usepackage{enumitem}
\usepackage{xcolor}
\usepackage{url}
\usepackage{balance}
\newcommand\crule[3][black]{\textcolor{#1}{\rule{#2}{#3}}}
\usepackage{booktabs, multirow} % for borders and merged ranges
\definecolor{roadcolor}{RGB}{234,51,246}
\definecolor{sidewalkcolor}{RGB}{68,8,72}
\definecolor{parkingcolor}{RGB}{241,156,249}
\definecolor{othergroundcolor}{RGB}{160,32,76}
\definecolor{buildingcolor}{RGB}{246,202,69}
\definecolor{carcolor}{RGB}{111,149,238}
\definecolor{truckcolor}{RGB}{74,32,172}
\definecolor{bicyclecolor}{RGB}{136,227,242}
\definecolor{motorcyclecolor}{RGB}{37,59,146}
\definecolor{othervehiclecolor}{RGB}{96,81,242}
\definecolor{vegetationcolor}{RGB}{79, 173, 50}
\definecolor{trunkcolor}{RGB}{126, 65, 22}
\definecolor{terraincolor}{RGB}{171, 238, 105}
\definecolor{personcolor}{RGB}{234, 60, 49}
\definecolor{bicyclistcolor}{RGB}{234, 66, 195}
\definecolor{motorcyclistcolor}{RGB}{138, 42, 90}
\definecolor{fencecolor}{RGB}{238, 128, 69}
\definecolor{polecolor}{RGB}{252, 241, 161}
\definecolor{trafficsigncolor}{RGB}{233, 51, 35}
\definecolor{color1}{RGB}{176, 36, 24}
\definecolor{color2}{RGB}{119,185,0}
\definecolor{color3}{RGB}{0, 0, 200}
\definecolor{colorofteaser}{RGB}{176, 36, 24}

\newcommand{\tbr}[1]{\textbf{\textcolor{color1}{#1}}}
\newcommand{\tbg}[1]{\textbf{\textcolor{color2}{#1}}}
\newcommand{\tbb}[1]{\textbf{\textcolor{color3}{#1}}}
\newcommand{\teaser}[1]{\textbf{\textcolor{colorofteaser}{#1}}}

\usepackage[pagebackref=true,breaklinks=true,letterpaper=true,colorlinks,bookmarks=false]{hyperref}
\hypersetup{citecolor=[RGB]{119,185,0}}
\usepackage[capitalize]{cleveref}
\crefname{section}{Sec.}{Secs.}
\Crefname{section}{Section}{Sections}
\Crefname{table}{Table}{Tables}
\crefname{table}{Tab.}{Tabs.}

\def\cvprPaperID{8506} % *** Enter the CVPR Paper ID here
\def\confName{CVPR}
\def\confYear{2023}

\begin{document}

% \title{Monocular 3D Semantic Scene Completion via Sparse 3D-to-2D Queries}

\title{VoxFormer: Sparse Voxel Transformer for  Camera-based\\ 3D Semantic Scene Completion}

\author{Yiming Li$^{1}$ \quad Zhiding Yu$^{2*}$ \quad  Christopher Choy$^2$ \quad Chaowei Xiao$^{2,3}$ \\ Jose M. Alvarez$^2$ \quad Sanja Fidler$^{2,4,5}$ \quad Chen Feng$^{1}$ \quad Anima Anandkumar$^{2,6}$\\
$^{1}$NYU \quad $^{2}$NVIDIA \quad $^{3}$ASU \quad $^{4}$University of Toronto \quad $^{5}$Vector Institute \quad $^{6}$Caltech
\\
% {\tt\small yimingli@nyu.edu, cfeng@nyu.edu}
% {\tt\small \url{github}}
% {\tt\small yimingli@nyu.edu, , cfeng@nyu.edu}
}

\maketitle

\renewcommand{\thefootnote}{\fnsymbol{footnote}} %将脚注符号设置为fnsymbol类型，即特殊符号表示
\footnotetext[1]{~Corresponding author: Zhiding Yu (zhidingy@nvidia.com)}
\footnotetext{}

%%%%%%%%% ABSTRACT
\begin{abstract}
\vspace{-2mm}
Humans can easily imagine the complete 3D geometry of occluded objects and scenes. This appealing ability is vital for recognition and understanding. To enable such capability in AI systems, we propose VoxFormer, a Transformer-based semantic scene completion framework that can output complete 3D volumetric semantics from only 2D images. Our framework adopts a two-stage design where we start from a sparse set of visible and occupied voxel queries from depth estimation, followed by a densification stage that generates dense 3D voxels from the sparse ones. A key idea of this design is that the visual features on 2D images correspond only to the visible scene structures rather than the occluded or empty spaces. Therefore, starting with the featurization and prediction of the visible structures is more reliable. Once we obtain the set of sparse queries, we apply a masked autoencoder design to propagate the information to all the voxels by self-attention. Experiments on SemanticKITTI show that VoxFormer outperforms the state of the art with a relative improvement of 20.0\% in geometry and 18.1\% in semantics and reduces GPU memory during training to less than 16GB. Our code is available on \url{https://github.com/NVlabs/VoxFormer}.

% despite possessing 752.3M fewer parameters.
\end{abstract}
\vspace{-6mm}

%%%%%%%%% BODY TEXT

\section{Introduction}
\label{sec:intro}
\vspace{-1mm}
Holistic 3D scene understanding is an important problem in autonomous vehicle (AV) perception. It directly affects downstream tasks such as planning and map construction. However, obtaining accurate and complete 3D information of the real world is difficult, since the task is challenged by the lack of sensing resolution and the incomplete observation due to the limited field of view and occlusions.
% , which poses challenges to holistic scene understanding.

To tackle the challenges, semantic scene completion (SSC)~\cite{song2017semantic} was proposed to jointly infer the complete scene geometry and semantics from limited observations. 
An SSC solution has to simultaneously address two subtasks: \textit{scene reconstruction} for visible areas and \textit{scene hallucination} for occluded regions. This task is further backed by the fact that humans can naturally reason about scene geometry and semantics from partial observations. However, there is still a significant performance gap between state-of-the-art SSC methods~\cite{roldao20223d} and human perception in driving scenes.

\begin{figure}[t]
  \centering
   \includegraphics[width=0.97\linewidth]{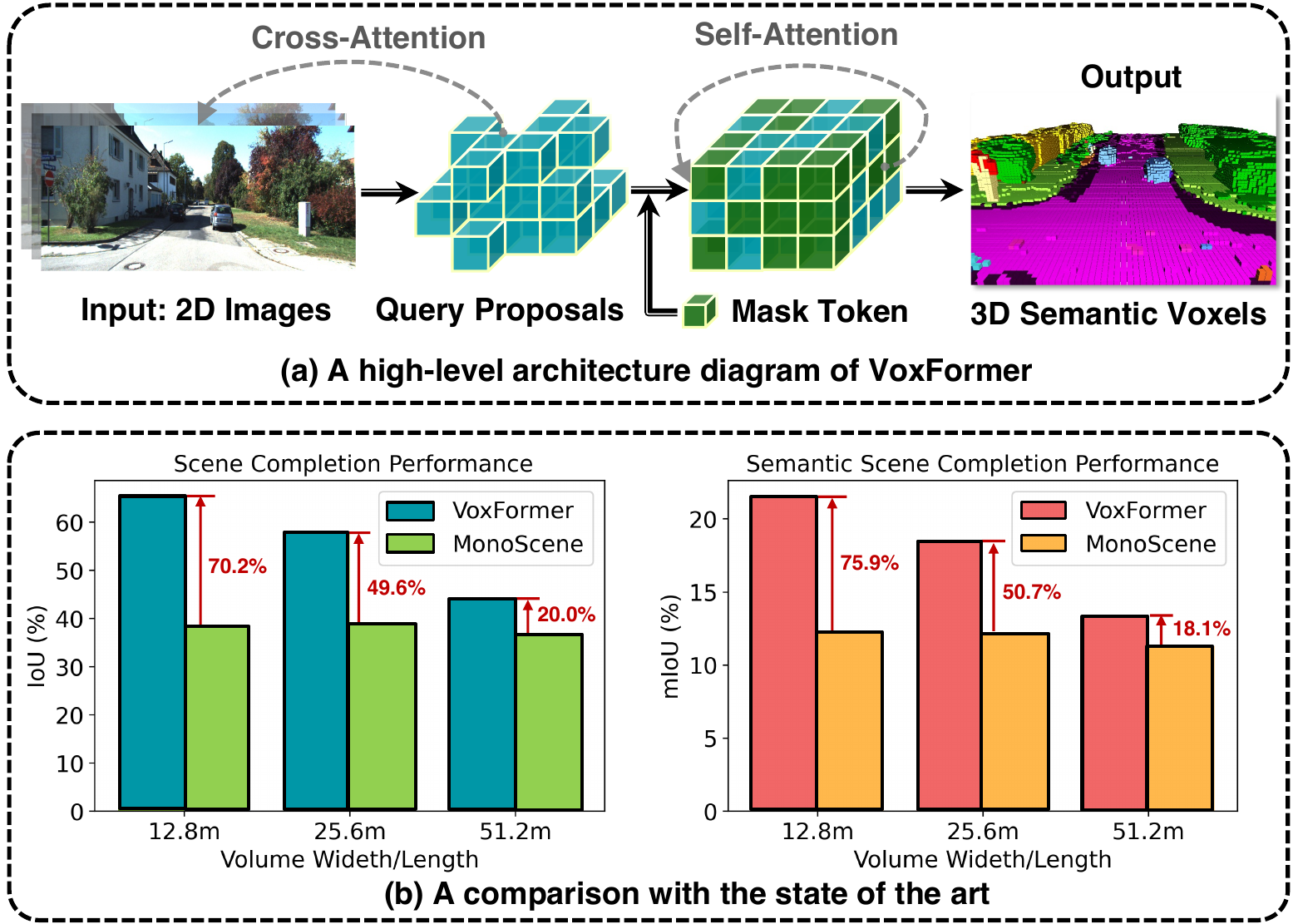}
   \caption{\textbf{(a) A diagram of VoxFormer} for camera-based semantic scene completion that predicts complete 3D geometry and semantics given only 2D images. After obtaining voxel query proposals based on depth, VoxFormer generates semantic voxels via an MAE-like architecture~\cite{he2022masked}. \textbf{(b) A comparison against the state-of-the-art MonoScene~\cite{cao2022monoscene}} in different ranges on SemanticKITTI~\cite{behley2019semantickitti}. VoxFormer performs much better in safety-critical short-range areas, while MonoScene performs indifferently at three distances. The relative gains are marked by \teaser{red}.}
   \label{fig:teaser}
   \vspace{-5mm}
\end{figure}

Most existing SSC solutions consider LiDAR a primary modality to enable accurate 3D geometric measurement~\cite{roldao2020lmscnet,cheng2021s3cnet,yan2021sparse,li2021semisupervised}. However, LiDAR sensors are expensive and less portable, while cameras are cheaper and provide richer visual cues of the driving scenes. This motivated the study of camera-based SSC solutions, as first proposed in the pioneering work of MonoScene~\cite{cao2022monoscene}. MonoScene lifts 2D image inputs to 3D using dense feature projection. However, such a projection inevitably assigns 2D features of visible regions to the empty or occluded voxels. For example, an empty voxel occluded by a car will still get the car's visual feature.
% Since each 2D pixel corresponds to multiple 3D voxels due to loss of depth, and the 2D-to-3D projection will associate a lot of empty voxels with false features, \eg, if an empty voxel is occluded by a car, it will be assigned the car's features since it shares an image pixel with a surface point from the car. 
As a result, the generated 3D features contain many ambiguities for subsequent geometric completion and semantic segmentation, resulting in unsatisfactory performance. 

% Besides, MonoScene requires many parameters ($\sim$1.8Gb) introduced by the 3D convolutional neural networks (CNNs).

\textbf{Our contributions.} Unlike MonoScene, VoxFormer considers 3D-to-2D cross-attention to represent the sparse queries. The proposed design is motivated by two insights: (1) \textit{reconstruction-before-hallucination}: the non-visible region's 3D information can be better completed using the reconstructed visible areas as starting points; and (2) \textit{sparsity-in-3D-space}: since a large volume of the 3D space is usually unoccupied, using a sparse representation instead of a dense one is certainly more efficient and scalable. Our contributions in this work can be summarized as follows:
% \vspace{-1mm}
\begin{itemize}[leftmargin=1.3em]
\item A novel two-stage framework that lifts images into a complete 3D voxelized semantic scene.
\item A novel query proposal network based on 2D convolutions that generates reliable queries from image depth.
\item A novel Transformer similar to masked autoencoder (MAE)~\cite{he2022masked} that yields complete 3D scene representation.
\item VoxFormer sets a new state-of-the-art in camera-based SSC on SemanticKITTI~\cite{behley2019semantickitti}, as shown in Fig.~\ref{fig:teaser} (b).
\end{itemize}

VoxFormer consists of \textit{class-agnostic query proposal} (stage-1) and \textit{class-specific semantic segmentation} (stage-2), where stage-1 proposes a sparse set of occupied voxels, and stage-2 completes the scene representations starting from the proposals given by stage-1. Specifically, stage-1 has a lightweight 2D CNN-based query proposal network using the image depth to reconstruct the scene geometry. It then proposes a sparse set of voxels from predefined learnable voxel queries over the entire field of view. Stage-2 is based on a novel sparse-to-dense MAE-like architecture as shown in Fig.~\ref{fig:teaser} (a). It first strengthens the featurization of the proposed voxels by allowing them to attend to the image observations. Next, the non-proposed voxels will be associated with a learnable mask token, and the full set of voxels will be processed by self-attention to complete the scene representations for per-voxel semantic segmentation.

Extensive tests on the large-scale SemanticKITTI~\cite{behley2019semantickitti} show that VoxFormer achieves state-of-the-art performance in geometric completion and semantic segmentation. More importantly, the improvements are significant in safety-critical short-range areas, as shown in Fig.~\ref{fig:teaser} (b).

\section{Related Works}
\textbf{3D reconstruction and completion.} \textit{3D reconstruction} aims to infer the 3D geometry of objects or scenes from single or multiple 2D images. This challenging problem receives extensive attention in the traditional computer vision era~\cite{hartley2003multiple} and the recent deep learning era~\cite{han2019image}. 3D reconstruction can be divided into (1) single-view reconstruction by learning shape priors from massive data~\cite{choy20163d,tulsiani2017multi,fan2017point,yan2016perspective}, and (2) multi-view reconstruction by leveraging different viewpoints~\cite{newcombe2011kinectfusion,ummenhofer2017demon}. Both explicit~\cite{choy20163d,fan2017point} and implicit representations~\cite{mescheder2019occupancy,peng2020convolutional,xu2019disn,popov2020corenet,zhu2022nice} are investigated for the object/scene. Unlike 3D reconstruction, \textit{3D completion} requires the model to hallucinate the unseen structure related to single-view 3D reconstruction, yet the input is in 3D except for 2D. 3D object shape completion is an active research topic that estimates the complete geometry from a partial shape in the format of point~\cite{yuan2018pcn,gu2020weakly,yan2022shapeformer}, voxels~\cite{chibane2020implicit,wang2021voxel,zhou20213d}, and distance fields~\cite{dai2017shape}, \etc. In addition to object-level completion, scene-level 3D completion has also been investigated in both indoor~\cite{dai2020sg} and outdoor scenes~\cite{vizzo2022make}: \cite{dai2020sg} proposes a sparse generative network to convert a partial RGB-D scan into a high-resolution 3D reconstruction with missing geometry. \cite{vizzo2022make} learns a neural network to convert each scan to a dense volume of truncated signed distance fields (TSDF).

\textbf{Semantic segmentation.} Human-level scene understanding for intelligent agents is typically advanced by semantic segmentation on images~\cite{9356353} or point clouds~\cite{guo2020deep}. Researchers have significantly promoted image segmentation performance with a variety of deep learning techniques, such as convolutional neural network~\cite{long2015fully,Noh_2015_ICCV}, vision transformers~\cite{xie2021segformer,cheng2021per}, prototype learning~\cite{wang2019panet,zhou2022rethinking}, \etc. To have intelligent agents interact with the 3D environment, thinking in 3D is essential because the physical world is not 2D but rather 3D. Thus various 3D point cloud segmentation methods have been developed to address 3D semantic understanding~\cite{qi2017pointnet,xu2020weakly,hu2020randla}. However, real-world sensing in 3D is inherently sparse and incomplete. For holistic semantic understanding, it is insufficient to solely parse the sparse measurements while ignoring the unobserved scene structures.

\textbf{3D semantic scene completion.} Holistic 3D scene understanding is challenged by limited sensing range, and researchers have proposed multi-agent collaborative perception to introduce more observations of the 3D scene~\cite{wang2020v2vnet, li2022v2x, xu2022v2x, li2021learning, xu2022cobevt, li2022multi, su2022uncertainty}.  Another line of research is 3D semantic scene completion, unifying scene completion and semantic segmentation which are investigated separately at the early stage~\cite{gupta2013perceptual, 1544938}. SSCNet~\cite{song2017semantic} first defines the semantic scene completion task where geometry and semantics are jointly inferred given an incomplete visual observation. In recent years, SSC in the indoor scenes with a relatively small scale has been intensively studied~\cite{zhang2018efficient,liu2018see,li2019depth,li2019rgbd,zhang2019cascaded,li2020anisotropic,chen20203d,cai2021semantic}. Meanwhile, SSC in the large-scale outdoor scenes have also started to receive attention after the release of SemanticKITTI dataset~\cite{behley2019semantickitti}. Semantic scene completion with a sparse observation is a highly desirable capability for autonomous vehicles since it can generate a dense 3D voxelized semantic representation of the scene. Such representation can aid 3D semantic map construction of the static environment and help perceive dynamic objects. Unfortunately, SSC for large-scale driving scenes is only at the preliminary development and exploration stages. Existing works commonly depend on 3D input such as LiDAR point clouds~\cite{roldao2020lmscnet,cheng2021s3cnet,rist2021semantic,li2021semisupervised,yan2021sparse}. In contrast, the recent MonoScene~\cite{cao2022monoscene} has studied semantic scene completion from a monocular image. It proposes 2D-3D feature projections and uses successive 2D and 3D UNets to achieve camera-only 3D semantic scene completion. However, 2D-to-3D feature projection is prone to introduce false features for unoccupied 3D positions, and the heavy 3D convolution will  lower the system's efficiency.

\begin{figure*}[t]
  \centering
   \includegraphics[width=0.925\linewidth]{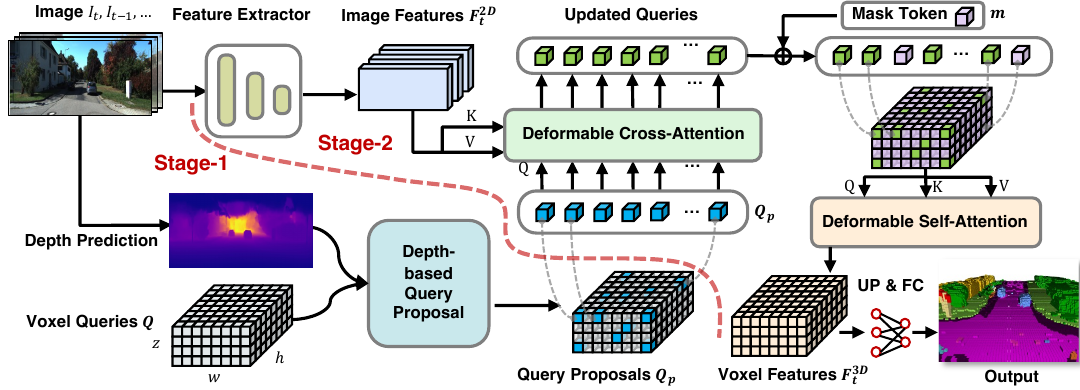}
   \caption{\textbf{Overall framework of VoxFormer.} Given RGB images, 2D features are extracted by ResNet50~\cite{he2016deep} and the depth is estimated by an off-the-shelf depth predictor. The estimated depth after correction enables the class-agnostic query proposal stage: the query located at an occupied position will be selected to carry out deformable cross-attention with image features. Afterwards, mask tokens will be added for completing voxel features by deformable self-attention. The refined voxel features will be upsampled and projected to the output space for per-voxel semantic segmentation. Note that our framework supports the input of single or multiple images.}
   \label{fig:main}
   \vspace{-2mm}
\end{figure*}

\textbf{Camera-based 3D perception.} Camera-based systems have received extensive attention in the autonomous driving community because the camera is low-cost, easy to deploy, and widely available. In addition, the camera can provide rich visual attributes of the scene to help vehicles achieve holistic scene understanding. Several works have recently been proposed for 3D object detection or map segmentation from RGB images. Inspired by DETR~\cite{carion2020end} in 2D detection, DETR3D~\cite{wang2022detr3d} links learnable 3D object queries with 2D images by camera projection matrices and enables end-to-end 3D bounding box prediction without non-maximum suppression (NMS). M2BEV~\cite{xie2022m} also investigates the viability of simultaneously running multi-tasks perception based on BEV features. BEVFormer~\cite{li2022bevformer} proposes a spatiotemporal transformer that aggregates BEV features from current and previous features via deformable attention~\cite{zhu2020deformable}. Compared to object detection, semantic scene completion can provide occupancy for each small cell instead of assigning a fixed-size bounding box to an object. This could help identify an irregularly-shaped object with an overhanging obstacle. Compared to 2D BEV representation, 3D voxelized scene representation has more information, which is particularly helpful when vehicles are driving over bumpy roads. Hence, dense volumetric semantics can provide a more comprehensive 3D scene representation, while how to create it with only cameras receives scarce attention.
\vspace{-1mm}

%-------------------------------------------------------------------------

\section{Methodology}
\subsection{Preliminary}\label{subsec:pre}
\vspace{-1mm}
\textbf{{Problem setup}.} We aim to predict a dense semantic scene within a certain volume in front of the vehicle, given only RGB images. More specifically, we use as input current and previous images denoted by $\mathbf{I}_t = \{ I_t, I_{t-1}, ...\}$, and use as output a voxel grid $\mathbf{Y}_t \in \{c_0, c_1, ..., c_M \}^{H \times W \times Z}$ defined in the coordinate of ego-vehicle at timestamp $t$, where each voxel is either empty (denoted by $c_0$) or occupied by a certain semantic class in $\{c_1, c_m, ..., c_M \}$. Here $M$ denotes the total number of interested classes, and $H$, $W$, $Z$ denote the length, width, and height of the voxel grid, respectively. In summary, the overall objective is to learn a neural network $\Theta$ to generate a semantic voxel $\mathbf{Y}_t = {\Theta}(\mathbf{I}_t)$ as close to the ground truth $\hat{\mathbf{Y}}_t$ as possible. Note that previous SSC works commonly consider 3D input~\cite{roldao20223d}. The most related work to us~\cite{cao2022monoscene} considers a single image as input which is our special case.

% \textbf{Design rationale.}
\textbf{Design rationale.} Motivated by \textit{reconstruction-before-hallucination} and \textit{sparsity-in-3D-space}, we build a two-stage framework: stage-1 based on CNN proposes a sparse set of voxel queries from image depth to attend to images since the image features correspond to visible and occupied voxels instead of non-visible and empty ones; stage-2 based on Transformer uses an MAE-like architecture to first strengthen the featurization of the proposed voxels by voxel-to-image cross-attention, and then process the full set of voxels with self-attention to enable the voxel interactions.

\subsection{Overall Architecture} 
We learn 3D voxel features from 2D images for SSC based on Transformer, as illustrated in Fig.~\ref{fig:main}: our architecture extracts 2D features from RGB images and then uses a sparse set of 3D voxel queries to index into these 2D features, linking 3D positions to an image stream using camera projection matrices. Specifically, voxel queries are 3D-grid-shaped learnable parameters designed to query features inside the 3D volume from images via attention mechanisms~\cite{vaswani2017attention}. Our framework is a two-stage cascade composed of class-agnostic proposals and class-specific segmentation similar to~\cite{ren2015faster}: stage-1 generates class-agnostic query proposals, and stage-2 uses an MAE-like architecture to propagate information to all voxels. Ultimately, the voxel features will be up-sampled for semantic segmentation. A more specific procedure is as follows:
\begin{itemize}[leftmargin=1.3em]
    \item Extract 2D features $\mathbf{F}^{2D}_t \in \mathbb{R}^{b \times c \times d}$  from RGB image $I_t$ using ResNet-50 backbone~\cite{he2016deep}, where $b\times c$ is spatial resolution, and $d$ is feature dimension.
    \item Generate class-agnostic query proposals $\mathbf{Q}_p \in \mathbb{R}^{N_p \times d}$ which is a subset of the predefined voxel queries $\mathbf{Q}\in \mathbb{R}^{N_q \times d}$, where $N_p$ and $N_q$ are the numbers of query proposals and the total number of voxel queries respectively. 
    
    \item Refine voxel features $\mathbf{F}^{3D}_t \in \mathbb{R}^{N_q \times d}$ with two steps: (1) update the subset of voxels corresponding to query proposals by using $\mathbf{Q}_p$ to attend to image features $\mathbf{F}^{2D}_t$ via cross-attention and (2) update all voxels by letting them attend to each other via self-attention. 

    \item Output dense semantic map $\mathbf{Y}_t \in \mathbb{R}^{H \times W \times Z \times (M+1)}$ by up-sampling and linear projection of $\mathbf{F}^{3D}_t$.
\end{itemize}
We will detail the voxel queries in~\cref{subsec:query}, stage-1 in~\cref{subsec:stage1}, stage-2 in~\cref{subsec:stage2}, and training loss in~\cref{subsec:train}.

\subsection{Predefined Parameters}\label{subsec:query}

\textbf{Voxel queries.} We pre-define a total of $N_q$ voxel queries as a cluster of 3D-grid-shaped learnable parameters $\mathbf{Q}\in \mathbb{R}^{h \times w \times z \times d}$ ($N_q = h \times w \times z$) as shown in the bottom left corner of Fig.~\ref{fig:main}, with $h \times w \times z$ its spatial resolution which is lower than output resolution $H \times W \times Z$ to save computations. Note that $d$ denotes the feature dimension, which is equal to that of image features. More specifically, a single voxel query $\mathbf{q} \in \mathbb{R}^d$ located at $(i,j,k)$ position of $\mathbf{Q}$ is responsible for the corresponding 3D voxel inside the volume. Each voxel corresponds to a real-world size of $a$ meters. Meanwhile, the voxel queries are defined in the ego vehicle's coordinate, and learnable positional embeddings will be added to voxel queries for attention stages, following existing works for 2D BEV feature learning~\cite{li2022bevformer}.
% Following common practices [14], we add learnable positional embedding to BEV queries Q before inputting them to VoxFormer.

\textbf{Mask token.} While some voxel queries are selected to attend to images; the remaining voxels will be associated with another learnable parameter to complete 3D voxel features. We name such learnable parameter as \textit{mask token}~\cite{he2022masked} for conciseness since unselected from $\mathbf{Q}$ is analogous to masked from $\mathbf{Q}$. Specifically, each mask token $\mathbf{m} \in \mathbb{R}^d$ is a learnable vector that indicates the presence of a missing voxel to be predicted. The positional embeddings are also added to help mask tokens be aware of their 3D locations. 

% \subsubsection{Query Proposal Stage}
\subsection{Stage-1: Class-Agnostic Query Proposal}\label{subsec:stage1}
Our stage-1 determines which voxels to be queried based on depth: the occupied voxels deserve careful attention, while the empty ones can be detached from the group. Given a 2D RGB observation, we first obtain a 2.5D representation of the scene based on depth estimation. Afterwards, we acquire 3D query positions by occupancy prediction that help correct the inaccurate image depth.

\textbf{Depth estimation.} We leverage off-the-shelf depth estimation models such as monocular depth~\cite{bhat2021adabins} or stereo depth~\cite{shamsafar2022mobilestereonet} to directly predict the depth $Z(u, v)$ of each image pixel $(u, v)$. Afterwards, the depth map $Z$ will be back-projected into a 3D point cloud: a pixel $(u, v)$ will be transformed to $(x, y, z)$ in 3D by:
\begin{equation}
    x=\frac{(u-c_u)\times z}{f_u}, y= \frac{(v-c_v)\times z}{f_v}, z=Z(u,v),
\end{equation}
where $(c_u , c_v)$ is the camera center and $f_u$ and $f_v$ are the horizontal and vertical focal length. However, the resulting 3D point cloud has low quality, especially in the long-range area, because the depth at the horizon is extremely inconsistent; only a few pixels determine the depth of a large area. 

\textbf{Depth correction.} To obtain satisfactory query proposals, we employ a model $\Theta_{occ}$ to predict an occupancy map at a lower spatial resolution to help correct the image depth. Specifically, the synthetic point cloud is firstly converted into a binary voxel grid map $\mathbf{M}_{in}$, where each voxel is marked as $1$ if occupied by at least one point. Then we can predict the occupancy by $\mathbf{M}_{out} = \Theta_{occ}(\mathbf{M}_{in})$, where $\mathbf{M}_{out} \in \{0, 1\}^{h \times w \times z}$ has a lower resolution than the input $\mathbf{M}_{in} \in \{0, 1\}^{H \times W \times Z}$ since a lower resolution is more robust to depth errors and compatible with the resolution of voxel queries. 
$\Theta_{occ}$ is a lightweight UNet-like model adapted from~\cite{roldao2020lmscnet}, mainly using 2D convolutions for binary classification of each voxel. 

\textbf{Query proposal.} Following depth correction, we can select voxel queries from $\mathbf{Q}$ based on the binary $\mathbf{M}_{out}$:
\begin{equation}
    \mathbf{Q}_p = \texttt{Reshape} (\mathbf{Q}[\mathbf{M}_{out}]),
\end{equation}
where $\mathbf{Q}_p \in \mathbb{R}^{N_p \times d}$ is the query proposals to attend to images later on. Our depth-based query proposal can: (1) save computations and memories by removing many empty spaces and (2) ease attention learning by reducing ambiguities caused by erroneous 2D-to-3D correspondences. 

\subsection{Stage-2: Class-Specific Segmentation}\label{subsec:stage2}
Following stage-1, we then attend to image features with query proposals $\mathbf{Q}_p$ to learn rich visual features of the 3D scene. For efficiency, we utilize deformable attention~\cite{zhu2020deformable}, which interacts with local regions of interest, and only sample $N_s$ points around the reference point  to compute the attention results. Mathematically, each query $\mathbf{q}$ will be updated by the following general equation:
\begin{equation}
    \texttt{DA}(\mathbf{q}, \mathbf{p}, \mathbf{F}) = \sum_{s=1}^{N_s} \mathbf{A}_{s} \mathbf{W}_s \mathbf{F}(\mathbf{p}+\delta \mathbf{p}_{s}),
\end{equation}
where $\mathbf{p}$ denotes the reference point, $\mathbf{F}$ represents input features, and $s$ indexes the sampled point from a total of $N_s$ points. $\mathbf{W}_s \in \mathbb{R}^{d \times d}$ denotes learnable weights for the value generation, $\mathbf{A}_{s} \in [0,1]$ is the learnable attention weight. $\delta \mathbf{p}_{s} \in \mathbb{R}^{2}$ is the predicted offset to the reference point $\mathbf{p}$, and $\mathbf{F}(\mathbf{p}+\delta \mathbf{p}_{s})$ is the feature at location $\mathbf{p}+\delta \mathbf{p}_{s}$ extracted by bilinear interpolation. Note that we only show the formulation of single-head attention for conciseness.

\textbf{Deformable cross-attention.} 
For each proposed query $\mathbf{q}_p$, we obtain its corresponding real-world location based on the voxel resolution $h\times w\times z$ and the real size of the interested 3D volume. Afterwards, we project the 3D point to 2D image features $\mathbf{F}^{2D} = \{ \mathbf{F}^{2D}_t, \mathbf{F}^{2D}_{t-1}, ...\}$ based on projection matrics. However, the projected 2D point can only fall on some images due to the limited field of view. Here, we term the hit image as $\mathcal{V}_{t}$. After that, we regard these 2D points as the reference points of the query
$\mathbf{q}_p$ and sample the features from the hit views around these reference points. Finally, we perform
a weighted sum of the sampled features as the output of deformable cross-attention (DCA):
\begin{equation}
\small
    \texttt{DCA}(\mathbf{q}_p, \mathbf{F}^{2D} ) = \frac{1}{|\mathcal{V}_{t}|} \sum_{t \in \mathcal{V}_{t}} \texttt{DA}(\mathbf{q}_p, \mathcal{P}(\mathbf{p},t), \mathbf{F}_t^{2D}),
\end{equation}
where $t$ indexes the images, and for each query proposal $\mathbf{q}_p$ located at $\mathbf{p}=(x,y,z)$, we use camera projection function $\mathcal{P}(\mathbf{p},t)$ to obtain the reference point on image $t$. 

\textbf{Deformable self-attention.}
After several layers of deformable cross-attention, the query proposals will be updated to $\hat{\mathbf{Q}}_p$. To get the complete voxel features, we combine the updated query proposals $\hat{\mathbf{Q}}_p$ and the mask tokens $\mathbf{m}$ to get the initial voxel features $\mathbf{F}^{3D} \in \mathbb{R}^{\times h\times w\times z \times d}$. Then we use deformable self-attention to get the refined voxel features $\hat{\mathbf{F}}^{3D}\in \mathbb{R}^{\times h\times w\times z \times d}$:
\begin{equation}
\small
    \texttt{DSA}(\mathbf{F}^{3D} , \mathbf{F}^{3D} ) = \texttt{DA}(\mathbf{f}, \mathbf{p}, \mathbf{F}^{3D}),
\end{equation}
where $\mathbf{f}$ could be either a mask token or an updated query proposal located at $\mathbf{p}=(x,y,z)$.

\textbf{Output Stage.} After obtaining refined voxel features $\hat{\mathbf{F}}^{3D}$, it will be upsampled and projected to the output space to get the final output
$\mathbf{Y}_t \in \mathbb{R}^{H \times W \times Z \times (M+1)}$, where $M+1$ denotes $M$ semantic classes and one empty class.

\subsection{Training Loss}\label{subsec:train}
% \textbf{Training objective.}
We train stage-2 with a weighted cross-entropy loss. The ground truth $\hat{\mathbf{Y}}_t \in \{c_0, c_1, ..., c_M \}^{H \times W \times Z}$ defined at time $t$ represents a multi-class semantic voxel grid. Therefore, the loss can be computed by:
\begin{equation}\label{eq:csc}
    \mathcal{L} = - \sum_{k=1}^{K} \sum_{c=c_0}^{c_M} w_c {\hat{y}}_{k,c} log(\frac{e^{{y}_{k,c}}}{\sum_c e^{{y}_{k,c}}}) ,
\end{equation}
where $k$ is the voxel index, $K$ is the total number of the voxel ($K = H \times W \times C$), $c$ indexes class, ${y}_{k,c}$ is the predicted logits for the $k$-th voxel belonging to class $c$,  $\hat{y}_{k,c}$ is the $k$-th element of ${\hat{\mathbf{{Y}}}}_t$ and is a one-hot vector (${y}_{i,k,c} = 1$ if voxel $k$ belongs to class $c$). $w_c$ is a weight for each class according to the inverse of the class frequency as in~\cite{roldao2020lmscnet}. We also use  scene-class affinity loss proposed in~\cite{cao2022monoscene}. For stage-1, we employ a binary cross-entropy loss for occupancy prediction at a lower spatial resolution.

\section{Experiments}
\begin{table*}[!htp]\centering
\renewcommand\tabcolsep{5pt}
\scriptsize
\begin{tabular}{l|ccc|ccc|ccc|ccc|cccc}\toprule
\textbf{Methods} &\multicolumn{3}{c|}{\textbf{VoxFormer-T (Ours)}} &\multicolumn{3}{c|}{\textbf{VoxFormer-S (Ours)}} &\multicolumn{3}{c|}{\textbf{MonoScene~\cite{cao2022monoscene}}} &\multicolumn{3}{c|}{\textbf{LMSCNet$^*$~\cite{roldao2020lmscnet}}} &\multicolumn{3}{c}{\textbf{SSCNet$^*$~\cite{song2017semantic}}} \\\midrule
\textbf{Range} &\textbf{12.8m} &\textbf{25.6m} &\textbf{51.2m} &\textbf{12.8m} &\textbf{25.6m} &\textbf{51.2m}  &\textbf{12.8m} &\textbf{25.6m} &\textbf{51.2m}  &\textbf{12.8m} &\textbf{25.6m} &\textbf{51.2m}  &\textbf{12.8m} &\textbf{25.6m} &\textbf{51.2m}  \\\midrule
\textbf{IoU (\%)} & {\tbg{65.38}} &{\tbr{57.69}} &{\tbr{44.15}} &{\tbb{65.35}} &{\tbg{57.54}} &{\tbg{44.02}} &38.42 &38.55 &36.80 &{\tbr{65.52}}  &{\tbb{54.89}} &38.36 &59.51 &53.20 &{\tbb{40.93}}  \\
\textbf{Precision (\%)} &\tbb{76.54} &\tbb{69.95} &\tbb{62.06} &\tbg{77.65} &\tbg{70.85} &\tbg{62.32} &51.22 &51.96 &52.19 &\tbr{86.51} &\tbr{82.21} &\tbr{77.60} &65.38 &59.13 &48.77 \\
\textbf{Recall (\%)} &\tbg{81.77} &\tbg{76.70} &\tbg{60.47} &\tbb{80.49} &\tbb{75.39} &\tbb{59.99} &60.60 &59.91 &55.50 &72.98 &62.29 &43.13 &\tbr{86.89} &\tbr{84.15} &\tbr{71.80} \\ \midrule
\textbf{mIoU} &\tbr{21.55} &\tbr{18.42} &\tbr{13.35} &{\tbg{17.66}} &{\tbg{16.48}} &{\tbg{12.35}} &12.25 &12.22 &{\tbb{11.30}} &{{15.69}} &14.13 &9.94 & \tbb{16.32} & {\tbb{14.55}} &10.27 \\
\crule[carcolor]{0.13cm}{0.13cm} \textbf{car} (3.92\%) &\tbr{44.90} &\tbr{37.46} &\tbr{26.54} &\tbb{39.78} &\tbb{35.24} &\tbg{25.79} &24.34 &24.64 &{23.29} &\tbg{42.99} &\tbg{35.41} & \tbb{23.62} &37.48 &31.09 &22.32 \\
\crule[bicyclecolor]{0.13cm}{0.13cm} \textbf{bicycle} (0.03\%) &\tbr{5.22} &\tbr{2.87} &\tbr{1.28} &\tbg{3.04} &\tbg{1.48} &\tbg{0.59} &\tbb{0.07} &\tbb{0.23} &\tbb{0.28} &0.00 &0.00 &0.00 &0.00 &0.00 &0.00 \\
\crule[motorcyclecolor]{0.13cm}{0.13cm} \textbf{motorcycle} (0.03\%)&\tbr{2.98} &\tbr{1.24} &\tbg{0.56} &\tbg{2.84} &\tbg{1.10} &\tbb{0.51} &\tbb{0.05} &\tbb{0.20} &\tbr{0.59} &0.00 &0.00 &0.00 &0.00 &0.00 &0.00 \\
\crule[truckcolor]{0.13cm}{0.13cm} \textbf{truck} (0.16\%) &\tbb{9.80} &\tbg{10.38} &\tbg{7.26} & 7.50 &{7.47} &\tbb{5.63} &\tbr{15.44} &\tbr{13.84} &\tbr{9.29} &0.76 &3.49 &1.69 &\tbg{10.23} & \tbb{8.49} &4.69 \\
\crule[othervehiclecolor]{0.13cm}{0.13cm} \textbf{other-veh.} (0.20\%) &\tbr{17.21} &\tbr{10.61} &\tbr{7.81} &\tbg{8.71} &\tbg{4.98} &\tbg{3.77} &\tbb{1.18} &\tbb{2.13} &\tbb{2.63} &0.00 &0.00 &0.00 &7.60 &4.55 &2.43 \\
\crule[personcolor]{0.13cm}{0.13cm} \textbf{person} (0.07\%) &\tbr{4.44} &\tbr{3.50} &\tbg{1.93} &\tbg{4.10} &\tbg{3.31} &\tbb{1.78} &\tbb{0.90} &\tbb{1.37} &\tbr{2.00} &0.00 &0.00 &0.00 &0.00 &0.00 &0.00 \\
\crule[bicyclistcolor]{0.13cm}{0.13cm} \textbf{bicyclist} (0.07\%)&\tbg{2.65} &\tbg{3.92} &\tbg{1.97} &\tbr{6.82} &\tbr{7.14} &\tbr{3.32} &\tbb{0.54} &\tbb{1.00} &\tbb{1.07} &0.00 &0.00 &0.00 &0.00 &0.02 &0.01 \\
\crule[motorcyclistcolor]{0.13cm}{0.13cm} \textbf{motorcyclist} (0.05\%) &0.00 &0.00 &0.00 &0.00 &0.00 &0.00 &0.00 &0.00 &0.00 &0.00 &0.00 &0.00 &0.00 &0.00 &0.00 \\
\crule[roadcolor]{0.13cm}{0.13cm} \textbf{road} (15.30\%) &\tbr{75.45} &\tbg{66.15} &{53.57} &\tbb{72.40} &{65.74} &\tbb{54.76} &57.37 &57.11 &\tbr{55.89} &\tbg{73.85} &\tbr{67.56} &\tbg{54.90} &72.27 & \tbb{65.78} &51.28 \\
\crule[parkingcolor]{0.13cm}{0.13cm} \textbf{parking} (1.12\%) &\tbr{21.01} &\tbr{23.96} &\tbr{19.69} &{10.79} &\tbb{18.49} &\tbg{15.50} &\tbg{20.04} &\tbg{18.60} &\tbb{14.75} & \tbb{15.63} &13.22 &9.89 &15.55 &13.35 &9.07 \\
\crule[sidewalkcolor]{0.13cm}{0.13cm} \textbf{sidewalk} (11.13\%) &\tbr{45.39} &\tbr{34.53} &\tbr{26.52} &{39.35} &\tbb{33.20} &\tbb{26.35} &27.81 &27.58 &\tbg{26.50} &\tbg{42.29} &\tbg{34.20} & {25.43} & \tbb{40.88} &32.84 &22.38 \\
\crule[othergroundcolor]{0.13cm}{0.13cm} \textbf{other-grnd}(0.56\%) &0.00 &\tbb{0.76} &\tbb{0.42} & {0.00} &\tbg{1.54} &\tbg{0.70} &\tbr{1.73} &\tbr{2.00} &\tbr{1.63} &0.00 &0.00 &0.00 &0.00 &0.01 &0.02 \\
\crule[buildingcolor]{0.13cm}{0.13cm} \textbf{building} (14.10\%) &\tbr{25.13} &\tbr{29.45} &\tbr{19.54} & {17.91} & {24.09} &\tbg{17.65} & {16.67} &15.97 &13.55 & \tbg{22.46} & \tbg{27.83} &14.55 & \tbb{18.19} &\tbb{24.59} &\tbb{15.20} \\
\crule[fencecolor]{0.13cm}{0.13cm} \textbf{fence} (3.90\%) &\tbr{16.17} &\tbr{11.15} &\tbg{7.31} &\tbg{12.98} &\tbg{10.63} &\tbr{7.64} &\tbb{7.57} &\tbb{7.37} &\tbb{6.60} &5.84 &4.42 &3.27 &5.31 &4.53 &3.57 \\
\crule[vegetationcolor]{0.13cm}{0.13cm} \textbf{vegetation} (39.3\%) &\tbr{43.55} &\tbr{38.07} &\tbr{26.10} &\tbg{40.50} &\tbg{34.68} &\tbg{24.39} &19.52 &19.68 &17.98 &\tbb{39.04} & \tbb{33.32} &20.19 & 36.34 & 33.17 &\tbb{22.24} \\
\crule[trunkcolor]{0.13cm}{0.13cm} \textbf{trunk} (0.51\%) &\tbr{21.39}  &\tbr{12.75} &\tbr{6.10} &\tbg{15.81} & \tbg{10.64} & \tbg{5.08} &2.02 &2.57 & 2.44 &6.32 &3.01 &1.06 &\tbb{13.35} & \tbb{8.53} & \tbb{4.33} \\
\crule[terraincolor]{0.13cm}{0.13cm} \textbf{terrain} (9.17\%) &\tbr{42.82} &\tbg{39.61} &\tbr{33.06} &32.25 & 35.08 &29.96 &31.72 &31.59 &29.84 &\tbg{41.59} &\tbr{41.51} & \tbg{32.30} &\tbb{37.61} &\tbb{38.46} &\tbb{31.21} \\
\crule[polecolor]{0.13cm}{0.13cm} \textbf{pole} (0.29\%) &\tbr{20.66} &\tbr{15.56} &\tbr{9.15} &\tbg{14.47} &\tbg{11.95} &\tbg{7.11} & 3.10 &3.79 &3.91 &7.28 &4.43 &2.04 &\tbb{11.36} &\tbb{8.33} & \tbb{4.83} \\
\crule[trafficsigncolor]{0.13cm}{0.13cm} \textbf{traf.-sign} (0.08\%) &\tbr{10.63} &\tbr{8.09} &\tbr{4.94} &\tbg{6.19} &\tbg{6.29} &\tbg{4.18} &3.69 & 2.54 &\tbb{2.43} &0.00 &0.00 &0.00 &\tbb{3.86} &\tbb{2.65} &1.49 \\
\bottomrule
\end{tabular}
\caption{\textbf{Quantitative comparison} against the state-of-the-art \textbf{camera-based} SSC methods. We report the performances inside three volumes, \ie, 12.8$\times$12.8$\times$6.4m$^3$, 25.6$\times$25.6$\times$6.4m$^3$, and 51.2$\times$51.2$\times$6.4m$^3$. The first two volumes are introduced for assessing the SSC performance in safety-critical nearby locations. The top three performances are marked by \tbr{red}, \tbg{green}, and \tbb{blue} respectively.}
\label{tab:comp2cam}
\vspace{-4mm}
\end{table*}

\subsection{Experimental Setup}
\textbf{Dataset.} We verify VoxFormer on SemanticKITTI~\cite{behley2019semantickitti}, which provides dense semantic annotations for each LiDAR sweep from the KITTI Odometry Benchmark~\cite{geiger2012we} composed of 22 outdoor driving scenarios. SemanticKITTI SSC benchmark is interested in a volume of $51.2m$ ahead of the car, $25.6m$
to left and right side, and $6.4m$ in height. The voxelization of this volume leads to a group of 3D voxel grids with a dimension of $256\times256\times32$ since each voxel has a size of
$0.2m \times 0.2m \times 0.2m$. The voxel grids are labelled with 20 classes (19 semantics and 1 free). Regarding the target output, SemanticKITTI provides the ground-truth semantic voxel grids by voxelization of the aggregated consecutive registered semantic point cloud. Regarding the sparse input to an SSC model, it can be either a single voxelized LiDAR sweep or an RGB image. In this work, we investigate image-based SSC similar to~\cite{cao2022monoscene}, yet our input could be multiple images including temporal information. 

\textbf{Implementation details.} Regarding stage-1, we employ the MobileStereoNet~\cite{shamsafar2022mobilestereonet} for direct depth estimation. Such depth can help generate a pseudo-LiDAR point cloud at a much lower cost based solely on stereo images. The occupancy prediction network for depth correction is adapted from LMSCNet~\cite{roldao2020lmscnet} which is on top of lightweight 2D CNNs. We directly utilize the depth predictor in~\cite{shamsafar2022mobilestereonet}, and we train an occupancy predictor from scratch, using as input a voxelized pseudo point cloud with a size of $256\times256\times32$ and as output an occupancy map with a size of $128\times128\times16$. Regarding stage-2, we crop RGB images of
cam2 to size $1220\times370$ and employ ResNet50~\cite{he2016deep} to extract image features, then the features in the 3rd stage will be taken by FPN~\cite{lin2017feature} to produce a feature map whose size is $1/16$ of the input image size. The feature dimension is set as $d=128$. The numbers of deformable attention layers for cross-attention and self-attention are $3$ and $2$ respectively. We use $8$ sampling points around each reference point for the cross-/self-attention head. There is also a linear layer that projects feature dimension $128$ to the number of classes $20$. 
% Note that the entire architecture composed of a depth estimator, an occupancy predictor, a ResNet50, an FPN, a cascade of deformable cross-/self-attention, and a linear projection has a total of $1.1$G parameters. 
We train stage-1 and stage-2 separately with $24$ epochs, a learning rate of $2\times10^{-4}$. Note that we provide two versions of VoxFormer, one takes only the current image as input (\textbf{VoxFormer-S}), and the other takes the current and the previous $4$ images as input (\textbf{VoxFormer-T}).

% \vspace{-0.5mm}
\textbf{Evaluation metrics.} We employ intersection over union (IoU) to evaluate the scene completion quality, regardless of the allocated semantic labels. Such a group of geometry-only voxel grids is actually a binary occupancy map which is crucial for obstacle avoidance in self-driving. We use the mean IoU (mIoU) of 19 semantic classes to assess the performance of semantic segmentation. Note that \textit{there is a strong interaction between IoU and
mIoU}, \eg, \textit{a high mIoU can be achieved by naively decreasing the IoU}. Therefore, the desired model should achieve excellent performance in both geometric completion and semantic segmentation. Meanwhile, we further propose to assess different ranges ahead of the car for a thorough evaluation: we individually report the IoU and mIoU inside the volume of $12.8m\times12.8m\times6.4m$, $25.6m\times25.6m\times6.4m$, and $51.2m\times51.2m\times6.4m$. Note that the understanding of a short-range area is more crucial since it leaves less time for autonomous vehicles to improve. Differently, the understanding of a \textit{provisional} long-range area could be enhanced as SDVs get closer to it to collect more observations. We report the results within different ranges on the validation set, and the results within the full range on the hidden test set are in the supplementary. 

\begin{figure*}[t]
  \centering
   \includegraphics[width=0.97\linewidth]{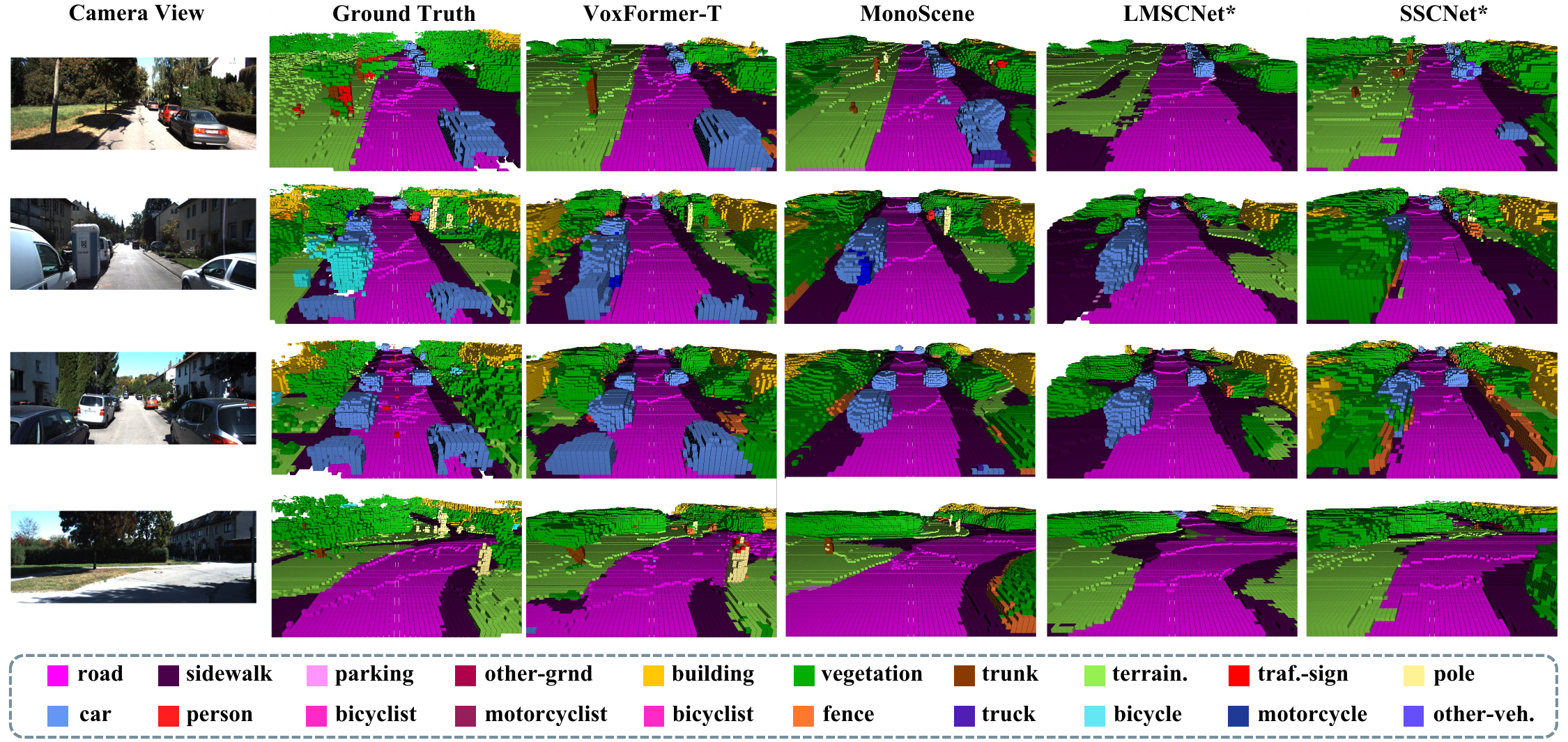}
   \vspace{-2mm}
   \caption{\textbf{Qualitative results of our method and others.} VoxFormer better captures the scene layout in large-scale self-driving scenarios. Meanwhile, VoxFormer shows satisfactory performances in completing small objects such as trunks and poles. }
   \label{fig:visualizations}
   \vspace{-3mm}
\end{figure*}

\textbf{Baseline methods.} We compare VoxFormer against the state-of-the-art SSC methods with public resources: (1) a camera-based SSC method MonoScene~\cite{cao2022monoscene} based on 2D-to-3D feature projection, (2) LiDAR-based SSC methods including JS3CNet~\cite{yan2021sparse}, LMSCNet~\cite{roldao2020lmscnet}, and SSCNet~\cite{song2017semantic}, and (3) RGB-inferred baselines LMSCNet$^*$~\cite{roldao2020lmscnet} and SSCNet$^*$~\cite{song2017semantic} which take as input a pseudo LiDAR point cloud generated by the stereo depth~\cite{shamsafar2022mobilestereonet}.

\subsection{Performance}
\subsubsection{Comparison against camera-based methods} 

\textbf{3D-to-2D query outperforms 2D-to-3D projection.} VoxFormer-S outperforms MonoScene by a large margin in terms of geometric completion ($36.80\rightarrow44.02, 19.62\%$); see Table~\ref{tab:comp2cam}. Such a large improvement stems from stage-1 with explicit depth estimation and correction, reducing a lot of empty spaces during the query process. In contrast, MonoScene based on 2D-to-3D projection will associate a lot of empty voxels with false features, \eg, if a free voxel is occluded by a car, it will be assigned the car's features when reprojected to the image, causing ambiguities during training.  Meanwhile, the semantic score  is also improved by $9.29\%$ without sacrificing IoU.

\begin{table}[t]\centering
\scriptsize
\renewcommand\tabcolsep{2pt}
\renewcommand\arraystretch{1.09}
\begin{tabular}{c|c|ccc|ccc}\toprule
\multirow{2}{*}{\textbf{Methods} }  & \multirow{2}{*}{\textbf{Modality} } &\multicolumn{3}{c|}{\textbf{IoU (\%)}} &\multicolumn{3}{c}{\textbf{mIoU (\%)}} \\
  &  &\textbf{12.8m} &\textbf{25.6m} & \textbf{51.2m} & \textbf{12.8m} & \textbf{25.6m} & \textbf{51.2m} \\\midrule
\textbf{MonoScene~\cite{cao2022monoscene}} & Camera & 38.42 & 38.55 & 36.80 & 12.25 & 12.22 & 11.30 \\
\textbf{VoxFormer-T (Ours)}& Camera & \tbg{65.38} &57.69 &44.15 & \tbb{21.55} &18.42 &13.35 \\\midrule
\textbf{SSCNet~\cite{song2017semantic}}& LiDAR &\tbb{64.37} & \tbb{61.02} &\tbb{50.22} &20.02 & \tbb{19.68} & \tbb{16.35} \\
\textbf{LMSCNet~\cite{roldao2020lmscnet}}& LiDAR &\tbr{74.88} & \tbr{69.45} & \tbr{55.22} & \tbg{22.37} & \tbg{21.50} & \tbg{17.19} \\
\textbf{JS3CNet~\cite{yan2021sparse}}& LiDAR &63.47 & \tbg{63.40} &\tbg{53.09} & \tbr{30.55} &\tbr{28.12} & \tbr{22.67} \\
\bottomrule
\end{tabular}
\vspace{-1mm}
\caption{\textbf{Quantitative comparison} against the state-of-the-art \textbf{LiDAR-based} SSC methods. VoxFormer even performs on par with some LiDAR-based methods at close range.}
\label{tab:comp2lidar}
\vspace{-5mm}
\end{table}

\textbf{Temporal information boost the semantic understanding.} Despite the negligible difference in IoU, VoxFormer-T further improves the SSC performance over VoxFormer-S with temporal information: the mIoU is improved by $8.10\%$, $11.77\%$, and $22.03\%$ inside three volumes ($51.2$m, $25.6$m, and $12.8$m) respectively as shown in Table~\ref{tab:comp2cam}. For example, the IoU scores of building, parking, and terrain categories are respectively improved by $10.71\%$, $27.03\%$, and $10.35\%$ inside the full volume because VoxFormer-S is restricted by the individual viewpoint while involving more viewpoints can mitigate this issue.

\textbf{Our superiority over others in short-range areas.} Our method shows a significant improvement over other camera-based methods in safety-critical short-range areas, as shown in Table~\ref{tab:comp2cam}. VoxFormer-T can achieve mIoU scores of $21.55$ and $18.42$ within $12.8$ meters and $25.6$ meters, which outperforms the state-of-the-art MonoScene by $75.92\%$ and $50.74\%$ respectively. Compared to MonoScene with comparable performances at different distances ($11.30\sim12.25$), VoxFormer with much better short-range performances is more desirable in self-driving. The reason is that the insufficient understanding of a provisional long-range area could be gradually advanced as SDVs move forward to collect more close observations. 

\begin{table}[t]\centering
 \renewcommand\tabcolsep{2pt}
\scriptsize
\begin{tabular}{c|c|ccc|ccc}\toprule
\multirow{2}{*}{\textbf{Methods}} &\multirow{2}{*}{\textbf{Depth}} &\multicolumn{3}{c|}{\textbf{IoU (\%)}} &\multicolumn{3}{c}{\textbf{mIoU (\%)}} \\
& &\textbf{12.8m} &\textbf{25.6m} & \textbf{51.2m} &\textbf{12.8m} &\textbf{25.6m} & \textbf{51.2m} \\ \midrule
\textbf{MonoScene~\cite{cao2022monoscene}} &-  & 38.42  & 38.55  & 36.80 & 12.25 &12.22  &\tbb{11.30}\\\midrule
\multirow{2}{*}{\textbf{VoxFormer-T (Ours)}} &Stereo~\cite{shamsafar2022mobilestereonet}  & \tbr{65.38}  & \tbr{57.69}  & \tbr{44.15} &\tbr{21.55}  & \tbr{18.42 } &  \tbr{13.35}\\
&Mono~\cite{bhat2021adabins} & \tbb{59.03} & 50.47 &  38.08 & \tbg{18.67} &\tbb{15.42} & 11.27\\ \midrule
\multirow{2}{*}{\textbf{VoxFormer-S (Ours)}} &Stereo~\cite{shamsafar2022mobilestereonet} & \tbg{65.35} &\tbg{57.54} & \tbg{44.02} & \tbb{17.66} &\tbg{16.48} & \tbg{12.35}\\
&Mono~\cite{bhat2021adabins} & 57.41 & \tbb{50.61}& \tbb{38.68} & 14.62 & 14.01& 10.67 \\
\bottomrule
\end{tabular}
\vspace{-1mm}
\caption{\textbf{Ablation study for image depth.} With monocular depth, VoxFormer-S performs better than MonoScene in geometry (12.8m, 25.6m, and 51.2m) and  semantics (12.8m and 25.6m). }
\label{tab:depth}
\vspace{-4mm}
\end{table}

% \begin{table}[t]\centering
% \scriptsize
% \renewcommand\arraystretch{1.18}
% \renewcommand\tabcolsep{1pt}
% \begin{tabular}{c|cc|cc|ccc}\toprule
% Range &\multicolumn{2}{c}{12.8m} &\multicolumn{2}{c}{25,6m} &\multicolumn{2}{c}{51.2m} \\\midrule
% Metrics &IoU (\%) &mIoU (\%) &IoU (\%) &mIoU (\%) &IoU (\%) &mIoU (\%) \\
% VoxFormer-T (Stereo) & & & & & & \\
% VoxFormer-T (Mono) & & & & & & \\
% VoxFormer-S (Stereo) & & & & & & \\
% VoxFormer-S (Mono) & & & & & & \\
% MonoScene & & & & & & \\
% \bottomrule
% \end{tabular}
% \caption{Generated by Spread-LaTeX}
% \label{tab:depth}
% \end{table}

\textbf{Our superiority over others for small objects.} VoxFormer shows a large advancement in completing small objects compared to the main baseline MonoScene such as the bicycle ($0.07\rightarrow5.22$), motorcycle ($0.05\rightarrow2.98$), bicyclist ($0.54\rightarrow6.82$), trunk ($2.02\rightarrow21.39$), pole ($3.10\rightarrow20.66$), and traffic sign ($3.69\rightarrow10.63$), as shown in Table \ref{tab:comp2cam}. The gap is even larger compared to LMSCNet$^*$ and SSCNet$^*$ directly consuming the pseudo point cloud, \eg, bicycle ($0.00\rightarrow5.22$), motorcycle ($0.00\rightarrow2.98$), and person ($0.00\rightarrow4.44$). Such major improvements come from the full exploitation of visual attributes of the 3D scene.

\textbf{Our superiority in size and memory.} VoxFormer has a total of $\sim$60M parameters, which is more lightweight than MonoScene with $\sim$150M parameters. Besides, VoxFormer needs less than 16GB GPU memory during training.

\begin{table}[t]\centering
\scriptsize
\renewcommand\tabcolsep{1.5pt}
\begin{tabular}{c|c|ccccccccc|c}\toprule
\textbf{Query} &\textbf{Dense}  &\multicolumn{9}{c|}{\textbf{Random}} & \textbf{Occupancy} \\\midrule
\textbf{Ratio (\%)}  &100 &90 &80 &70 &60 &50 &40 &30 &20 &10 & 10$\sim$20 \\\midrule
\textbf{Memory (G)} & {18.5} &18.2 &17.6 &17.3 &16.8 &16.3 &15.8 &15.3 &14.9 &\textbf{14.4}&14.6 \\
\textbf{IoU (\%)} &34.6 &34.5 &34.1 &34.0 &34.2 &33.9 &24.5 &34.0 &33.5 &24.6 & \textbf{44.0} \\
\textbf{mIoU (\%)} &10.1 &9.9 &9.9 &9.8 &9.6 &9.5 &3.8 &9.3 &8.9 &3.8 & \textbf{12.4} \\
\bottomrule
\end{tabular}
\vspace{-2mm}
\caption{\textbf{Ablation study for query proposal.} Our depth-based query proposal performs best.}
\label{tab:query}
\vspace{-2mm}
\end{table}

\begin{table}[t]\centering
\scriptsize
\renewcommand\tabcolsep{1pt}
\begin{tabular}{lcccccccccccc}\toprule
${t}$ &-10 &0 &+10 &+20 &+30 &+40 &+50 &+60 & \textbf{IoU (\%)} & \textbf{mIoU (\%)} & \textbf{Memory (G)} \\\midrule
\textbf{Online} &\checkmark &\checkmark & & & & & & &44.31 &13.24 &\textbf{15.21} \\\midrule
\multirow{6}{*}{\textbf{Offline}} &\checkmark &\checkmark &\checkmark & & & & & &44.48 &14.02 &15.74 \\
&\checkmark &\checkmark &\checkmark &\checkmark & & & & &44.24 &14.53 &16.25 \\
&\checkmark &\checkmark &\checkmark &\checkmark & \checkmark & & & &44.83 &15.42 &16.81 \\
&\checkmark &\checkmark &\checkmark &\checkmark &\checkmark &\checkmark & & &44.58 &15.88 &17.43 \\
&\checkmark &\checkmark &\checkmark &\checkmark &\checkmark &\checkmark & \checkmark & &44.53 &16.09 &18.03 \\
&\checkmark &\checkmark &\checkmark &\checkmark &\checkmark &\checkmark &\checkmark &\checkmark &\textbf{45.05} & \textbf{16.20} &19.37 \\
\bottomrule
\end{tabular}
\vspace{-2mm}
\caption{\textbf{Ablation study for temporal input.} $+N$ means using the future frame $t+N$. Memory denotes training memory.}\label{tab:video}
\vspace{-5mm}
\end{table}

\vspace{-3mm}
\subsubsection{Comparison against LiDAR-based methods} As shown in Table \ref{tab:comp2lidar}, as the distance gets closer to the ego-vehicle, the performance gap between our method and the state-of-the-art LiDAR-based methods becomes smaller, \eg, compared to LMSCNet, the mIoU pair is $13.35\leftrightarrow17.19$ if considering the area of $51.2\times51.2m^2$ ahead of the ego-vehicle, while the mIoU pair will be $21.55\leftrightarrow22.37$ if only considering the area of $12.8\times12.8m^2$. This observation is promising and inspiring to the self-driving community since VoxFormer only needs cheap cameras during inference. More interestingly, our mIoU within $12.8\times12.8m^2$ is even better than LiDAR-based SSCNet with a relative gain of $7.63\%$, and our IoU within $12.8\times12.8m^2$ is better than LiDAR-based JS3CNet with an improvement of $3.00\%$. In contrast, there is always a large gap between MonoScene and LiDAR-based methods at different ranges.

\vspace{-2mm}
\subsubsection{Ablation studies}
\textbf{Depth estimation.} We compare the performances between VoxFormers using monocular~\cite{bhat2021adabins} and stereo depth~\cite{shamsafar2022mobilestereonet}, as shown in Table~\ref{tab:depth}. In general, stereo depth is more accurate than monocular depth since the former exploits epipolar geometry, but the latter relies on pattern recognition~\cite{laga2020survey}. Hence, VoxFormer with stereo depth performs best. Note that our framework can be integrated with any state-of-the-art depth models, so using a stronger existing depth predictor~\cite{xie2022revealing,tankovich2021hitnet,yuan2022newcrfs} could enhance our SSC performance. Meanwhile, VoxFormer can be further promoted along with the advancement of depth estimation.

\textbf{Query mechanism.} The ablation study for the query mechanism is reported in Table~\ref{tab:query}. We find that: (1) dense query (use all voxel queries in stage-2) is inefficient in memory consumption and performs worse than our occupancy-based query in both geometry and semantics; (2) the performance of random query (randomly proposing a subset from all $128\times128\times16$ voxel queries based on a specific ratio) is not stable, and there is a large gap between the random and occupancy-based query in both geometry and semantics; (3) our method achieves an excellent tradeoff between the memory consumption and the performance.

\textbf{Temporal input.} The ablation study for temporal information is shown in Table~\ref{tab:video}. The offline setting with more future observations can largely boost semantic segmentation. Compared to the online setting with only previous and current images, the mIoU can be improved from $13.24$ to $16.20$ ($22.36\%$). Note that involving more temporal input can lead to more memory consumption.

\textbf{Image features.} The ablation study for 2D feature layers is shown in Table~\ref{tab:layer}. We see that using different layers has comparable IoU but different mIoU. Using the layer whose size is  1/16 of the input image size achieves an excellent balance between the performance and the model size.

\textbf{Architecture.} We conduct architecture ablation as shown in Table~\ref{tab:architecture}. For stage-1, depth estimation and correction are both important since a group of reasonable voxel queries can set a good basis for complete scene representation learning. For stage-2, self-attention and cross-attention can help improve the performance by enabling voxel-to-voxel and voxel-to-image interactions. 

\vspace{-2mm}
\subsubsection{Limitation and future work}
Our performance at long range still needs to be improved, because the depth is very unreliable at the corresponding locations. Decoupling the long-range and short-range SSC is a potential solution to enhance the SSC far away from the ego vehicle. We leave this as our future work.

\begin{table}[t]\centering
\scriptsize
\begin{tabular}{lccccccc}\toprule
\multicolumn{4}{c}{\textbf{Spatial resolution}} &\multirow{2}{*}{\textbf{IoU (\%)}} &\multirow{2}{*}{\textbf{mIoU (\%)}} &\multirow{2}{*}{\textbf{Params (M)}} \\
$\frac{1}{4}$ & $\frac{1}{8}$ &$\frac{1}{16}$ & $\frac{1}{32}$ & & \\ \midrule
\checkmark & & & & 44.26 & 10.24 &\textbf{57.81} \\
&\checkmark & & & 44.38 & 11.33 & 57.84 \\
& &\checkmark & &\textbf{44.02} &\textbf{12.35} & 57.90 \\
& & &\checkmark &44.19 &12.29 & 58.04 \\
\checkmark &\checkmark &\checkmark &\checkmark &44.01 &12.22 & 58.93 \\
\bottomrule
\end{tabular}
\vspace{-2mm}
\caption{\textbf{Ablation study for 2D image feature layers.} Spatial resolution is relative to the input image size.}
\label{tab:layer}
\vspace{-2mm}
\end{table}

\begin{table}[t]\centering
\scriptsize
\renewcommand\tabcolsep{12pt}
\begin{tabular}{cccc}\toprule
\textbf{Methods} & \textbf{IoU (\%)} & \textbf{mIoU (\%)} \\\midrule
\textbf{Ours} & \textbf{44.02}& \textbf{12.35} \\
\textbf{Ours w/o depth estimation} & 34.64 & 10.14 \\
\textbf{Ours w/o depth correction} & 36.95 & 11.36 \\
\textbf{Ours w/o cross-attention} & 32.74 & 9.94\\
\textbf{Ours w/o self-attention} & 43.73 & 10.70\\
\bottomrule
\end{tabular}
\vspace{-2mm}
\caption{\textbf{Ablation study for architecture.}}\label{tab:architecture}
\vspace{-5mm}
\end{table}

\section{Conclusion}
In this paper, we present VoxFormer, a strong camera-based 3D semantic scene completion (SSC) framework
composed of (1) class-agnostic query proposal based on depth estimation and (2) class-specific segmentation with a sparse-to-dense MAE-like design. VoxFormer outperforms the state-of-the-art camera-based method
and even performs on par with LiDAR-based methods at close range. We hope VoxFormer can motivate further research in camera-based SSC and its applications in AV perception. 

%------------------------------------------------------------------------
\renewcommand{\thetable}{\Roman{table}}
\renewcommand{\thefigure}{\Roman{figure}}
\renewcommand\thesection{\Alph {section}}

\section*{Appendix}
\setcounter{section}{0}
\setcounter{figure}{0}
\setcounter{table}{0}

\balance
In the appendix, we mainly provide quantitative and qualitative results of our method and the state-of-the-art camera-based SSC method MonoScene~\cite{cao2022monoscene} on the hidden test set of SemanticKITTI~\cite{behley2019semantickitti}. Since we do not have access to the ground truth of the test set, we can only report  the performances  within the full range (51.2$\times$51.2$\times$6.4m$^3$).

\section{Quantitative Comparison}
\textbf{Scene completion.} As shown in Table~\ref{tab:hiddentest}, VoxFormer outperforms MonoScene with a large gap in terms of geometric completion. VoxFormer-S without using historical observations improves MonoScene on IoU with a relative gain of $25.73\%$. Note that in autonomous driving, geometry occupancy is critical for obstacle avoidance since a false negative could result in severe accidents. Therefore, our method is more desirable than MonoScene in safety-critical camera-based autonomous driving applications. 

\textbf{Semantic scene completion.} As shown in Table~\ref{tab:hiddentest}, VoxFormer also demonstrates a better semantic scene understanding. VoxFormer-S and VoxFormer-T both demonstrate better mIoU than MonoScene. VoxFormer-T / VoxFormer-S have a relative improvement of $21.03\%$ / $10.11\%$ compared with the cutting-edged MonoScene. Note that the values of IoU and mIoU are intertwined, and some methods can naively increase the value of mIoU by sacrificing IoU. In contrast, our method shows  superior performance in terms of both geometry and semantics.

\textbf{Short-range performances.} Although short-range evaluations are not available on the hidden test set, we expect to see a similar trend (we perform much better in safety-critical short-range areas than MonoScene). The reason is that the scores of mIoU and IoU on the test set are comparable to that on the validation set inside the 51.2$\times$51.2$\times$6.4m$^3$ volume. For example, VoxFormer-S achieves an mIoU of 12.35 on the validation set and 12.20 on the test set.

\section{Qualitative Comparison}
\begin{figure*}[t]
  \centering
   \includegraphics[width=0.88\linewidth]{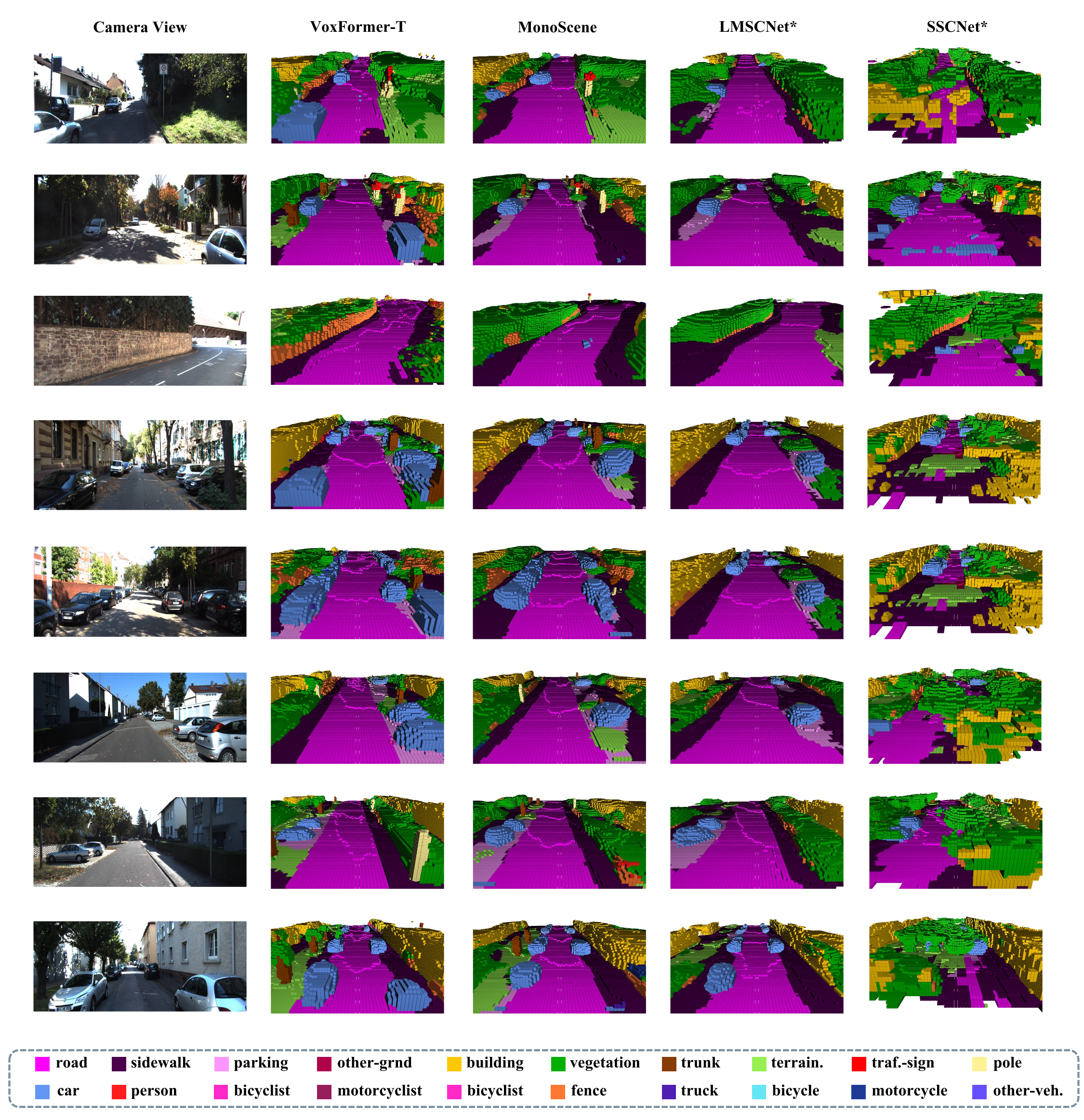}
   \vspace{-2mm}
   \caption{\textbf{Qualitative results of our method and others on the hidden test set.} VoxFormer better captures the scene layout in large-scale self-driving scenarios. Meanwhile, VoxFormer shows satisfactory performances in completing small objects such as trunks and poles. }
   \label{fig:appendix}
   \vspace{-3mm}
\end{figure*}

More visualizations are shown in \cref{fig:appendix}. We can see that our method performs much better than MonoScene in the short-range areas. There are some missing objects for MonoScene at close range, as shown in the first and last row of \cref{fig:appendix}. Meanwhile, the long-range performance of our method can be further improved, \eg, the trunks in the long-range areas are not completed in the fourth row of \cref{fig:appendix}.

\begin{table*}[htbp]
\renewcommand\tabcolsep{2.5pt}
\scriptsize
    \centering
    \begin{tabular}{c|c|c|c|c|c|c|c|c|c|c|c|c|c|c|c|c|c|c|c|c|cc}
   \toprule 
\rotatebox{90}{\textbf{Method}}&\rotatebox{90}{\textbf{IoU}}
& \rotatebox{90}{\crule[carcolor]{0.13cm}{0.13cm} \textbf{car} (3.92\%)}
& \rotatebox{90}{\crule[bicyclecolor]{0.13cm}{0.13cm} \textbf{bicycle} (0.03\%)}
& \rotatebox{90}{\crule[motorcyclecolor]{0.13cm}{0.13cm} \textbf{motorcycle} (0.03\%)} 
& \rotatebox{90}{\crule[truckcolor]{0.13cm}{0.13cm} \textbf{truck} (0.16\%)} 
& \rotatebox{90}{\crule[othervehiclecolor]{0.13cm}{0.13cm} \textbf{other-veh.}(0.20\%)}
& \rotatebox{90}{\crule[personcolor]{0.13cm}{0.13cm} \textbf{person} (0.07\%)}
& \rotatebox{90}{\crule[bicyclistcolor]{0.13cm}{0.13cm} \textbf{bicyclist} (0.07\%)}
& \rotatebox{90}{\crule[motorcyclistcolor]{0.13cm}{0.13cm} \textbf{motorcyclist} (0.05\%)}
& \rotatebox{90}{\crule[roadcolor]{0.13cm}{0.13cm} \textbf{road} (15.30\%)}
& \rotatebox{90}{\crule[parkingcolor]{0.13cm}{0.13cm} \textbf{parking} (1.12\%)} 
& \rotatebox{90}{\crule[sidewalkcolor]{0.13cm}{0.13cm} \textbf{sidewalk} (11.13\%)}
& \rotatebox{90}{\crule[othergroundcolor]{0.13cm}{0.13cm} \textbf{other-grnd}(0.56\%)}
& \rotatebox{90}{\crule[buildingcolor]{0.13cm}{0.13cm} \textbf{building} (14.10\%)}
& \rotatebox{90}{\crule[fencecolor]{0.13cm}{0.13cm} \textbf{fence} (3.90\%)}
& \rotatebox{90}{\crule[vegetationcolor]{0.13cm}{0.13cm} \textbf{vegetation} (39.3\%)}
& \rotatebox{90}{\crule[trunkcolor]{0.13cm}{0.13cm} \textbf{trunk} (0.51\%)}
&\rotatebox{90}{\crule[terraincolor]{0.13cm}{0.13cm} \textbf{terrain} (9.17\%)}
&\rotatebox{90}{\crule[polecolor]{0.13cm}{0.13cm} \textbf{pole} (0.29\%)} 
&\rotatebox{90}{\crule[trafficsigncolor]{0.13cm}{0.13cm} \textbf{traf.-sign} (0.08\%)}
&\rotatebox{90}{\textbf{mIoU}}
\\ \midrule
\textbf{MonoScene} & 34.16 & 18.80 & 0.50 & 0.70 & {3.30} & \textbf{4.40} & {1.00} & 1.40 & \textbf{0.40}& \textbf{54.70}& {24.80}& \textbf{27.10}& 5.70& 14.40& {11.10}& 14.90&2.40& 19.50& 3.30& 2.10& 11.08 \\\midrule
\textbf{VoxFormer-S (Ours)} & {42.95} & 20.80 & 1.00 & 0.70 & 3.50 & 3.70 & {1.40} & \textbf{2.60} & 0.20 & 53.90 & 21.10 & 25.30 &{5.60} & 19.80 & 11.10 &22.40 & 7.50& 21.30& 5.10 & 4.90 & 12.20\\
\textbf{VoxFormer-T (Ours)} & \textbf{43.21} & \textbf{21.70} & \textbf{1.90} & \textbf{1.60} & \textbf{3.60} & 4.10& \textbf{1.60} & {1.10}& 0.00& 54.10 & \textbf{25.10} & 26.90& \textbf{7.30} & \textbf{23.50}& \textbf{13.10}& \textbf{24.40}& \textbf{8.10}& \textbf{24.20}& \textbf{6.60}& \textbf{5.70}& \textbf{13.41}\\

\bottomrule
\end{tabular}
    \caption{Quantitative results of VoxFormer and the state-of-the-art MonoScene on the hidden test set of SemanticKITTI.}
    \label{tab:hiddentest}
\end{table*}

%%%%%%%% REFERENCES
{\small
\balance
\bibliographystyle{unsrt}
\bibliography{egbib}
}

\end{document}